\documentstyle[eaclap]{article}

\title{\vspace{-0.5in}Cooperative Error Handling and Shallow Processing}
\author{Tanya Bowden\\
Computer Laboratory\\
University of Cambridge\\
Pembroke St.\ \\
Cambridge  CB2 3QG \\
 U.\ K.\ \\
Tanya.Bowden@cl.cam.ac.uk\\}

\begin{document}

\maketitle
\vspace{-0.5in}
\begin{abstract}

This paper is concerned with the detection and correction of
sub-sentential English text errors.
Previous spelling programs, unless restricted to a very small set
of words, have operated as post-processors.  And to date,
grammar checkers and other programs which deal with ill-formed input
usually step directly from spelling considerations to a full-scale
parse, assuming a complete sentence.
Work described below is aimed at evaluating the effectiveness
of shallow (sub-sentential) processing and the feasibility of cooperative
error checking, through building and testing appropriately an
error-processing system.
A system under construction is outlined which incorporates
morphological checks (using new two-level error rules) over a directed letter
graph,
tag positional trigrams and partial parsing.  Intended testing is discussed.

\end{abstract}

Unless a keyboard user is particularly proficient, a frustrating amount of time
is
usually spent backtracking to pick up mis-typed or otherwise mistaken input.
Work described in this paper started from an idea of an error processor that
would sit on top of an editor, detecting/correcting errors just after entry,
while
the user continued with further text, relieved from tedious backtracking.
Hence `co-operative' error processing.  But if a program is to catch such
errors very soon after they are entered, it will have to operate with less
than the complete sentence.

Work underway focuses on shallow processing: how far error detection and
correction can proceed when the system purview is set to a stretch of text
which
does not admit complete sentential analysis.  To date, grammar checkers and
other programs which deal with illformed input usually step directly from
spelling considerations to a full-scale sentence parse.
However treating the sentence as a basic unit loses meaning when the
`sentence' is incomplete or illformed.
Shallow processing is also interesting because it should be cheaper and
faster than a complete analysis of the whole sentence.

To investigate issues involved in shallow processing and cooperative
error handling, the {\it pet} (processing errors in text) system is
being built.  The focus is on these two issues; no attempt is being
made to produce a complete product
\footnote{
In particular, there are many HCI issues associated with such a system,
which are beyond the scope of this paper.
}.
{\it Pet} operates over a shifting window of text
(it can be attached simply and asynchronously to the Emacs editor).  One
word in this purview is in focus at a time.  {\it Pet} will give
one of three responses to this word; it will accept the word, suggest a
correction,
or indicate that it found an error it couldn't correct.
Below follow an outline and discussion of the (linguistic) components of {\it
pet} and
discussion of testing and evaluation of the system.

\section*{Pet System}

\noindent
{\it Morphological Processing \& Spelling Checking}

\smallskip
The word in focus is first passed through a two-level morphological analysis
stage,
based on an adaption of (Pulman, 1991).  Two purposes are served here: checking
the word is lexical (i.e. in the lexicon or a permissible inflection of
a word in the lexicon) and collecting the possible categories, which
are represented as sets of feature specifications (Grover, 1993).

 This morphological lookup operates over a character trie which
has been compressed into a (directed) graph. Common endings are shared
and category information is stored on the first unique transition.  The
advantages of this compression are that (1) a word/morpheme is
recognised (and category affixation rules (Grover, 1993) checked) as soon as
the initial
letters allow uniqueness, rather than at the end of the word, and (2) there
is an immense saving of space. There was a reduction of over half the
transitions on
the trie formed from the Alvey lexicon.

If the word is unknown, the system reconsiders analysis from the point
where it broke down with the added possibility of an error rule.
There are currently four error rules, corresponding to the four Damerau
transformations: omission, insertion, transposition, substitution (Damerau,
1964) -
considered in that order (Pollock, 1983).
The error rules are in two level format and integrate seamlessly into
morphological
analysis.

\smallskip

\begin{center} \mbox{ * - X - * $\rightarrow$ * - - *} \end{center}

\small
 This says that any letter (`X') can be inserted, with asterisks
indicating that it can occur in any context (compare with (Pulman, 1991)).  The
right hand side represents the `error surface' and the left hand side
the surface with error removed.
\normalsize

\medskip

\noindent
If this doesn't succeed, it backtracks to try an error rule at an earlier point
in the analysis.  At present it will not apply more than one error rule per
word, in keeping with findings on error frequencies (Pollock, 1983).

As an alternative, a program was developed which uses positional
binary trigrams (Riseman,1974) (p.b.t.'s) to spot the error position and to
check candidate
corrections generated by reverse Damerau transformations.  This
should have the advantage over the two level error rules in that it
uses a good method of calculating likely error positions and because
a set of correction possibilities can be generated fairly cheaply.
(Correction possibilities are ranked using frequency information on Damerau
errors and by giving preference to very common words.)
However initial tests over a small file of constructed errors showed that
the error rules did just as well (slightly better in fact) at
choosing the `correct correction'.

The error rules are applied when ordinary
morphological rules fail - which is usually a place p.b.t.'s would
mark as in error - but the rules don't ignore error locations p.b.t.'s
accept as allowable letter combinations.  Most importantly, the
error rules operate over a letter graph of the lexicon, so only ever consider
lexical words (unknown letters are instantiated to the letters
associated with the transition options).  The disadvantage remains
that generating many correction possibilities (with SICStus
backtracking) is time-consuming.  At present this phase postulates only
one grapheme at a time, although all its possible categories are passed along
together to later stages.  If all of these categories eventually fail analysis,
backtracking to alternative correction candidates (different graphemes)
will occur.

\bigskip
\noindent
{\it Tag Checking  \& Partial Parsing}

\smallskip
The Alvey features are mapped on to the CLAWS tagset used in the LOB corpus
(Garside, 1987).
Tag transitions are checked against an occurrence matrix of the tagged LOB
corpus
using positional binary trigrams similar to those used in the spelling checks
mentioned above.  Tag checks though the current set of categories stop when one
category passes, but backtrack and continue if parsing then fails.

The Core Language Engine (CLE) is an application independent, unification based
``general purpose device for mapping between natural language sentences and
logical form representations" (Alshawi, 1992).  Its intermediate syntactic
stages involve phrasal parsing followed by full syntactic analysis (top-down,
left-corner).  If the latter stage fails, CLE invokes partial parsing.

The phrasal phase and partial parsing have been extracted and are
being adapted to the present purpose.  After mapping onto CLE tags,
application of the phrasal phase, which implements bottom-up parsing, is
straightforward.  CLE partial parsing, using left-corner analysis
combined with top-down prediction on the results of the phrasal phase,
looks for complete phrases and breaks down a wordstring into
maximal segments.

\small
\begin{tabbing}
(a)aa\= \kill
(a) the the brown bear $\rightarrow$ the $\mid$ the brown bear \\
(b) ate the nice friendly $\rightarrow$ ate $\mid$ the $\mid$ nice $\mid$
friendly
\end {tabbing}

\normalsize
\noindent
For example, (a) produces 1 segment and (b) produces 4 segments -
whereas ``ate the nice friendly cat" would produce 1 segment.

Partial parsing needs to be adapted to support the idea of the {\it pet}
purview; partial parsing that accepts any string likely to constitute
part of a sentence.  To achieve this the ends of the wordstring delimited
by the purview need to be treated differently.  On the right hand end,
`can start rule' possibilities of words can be considered, using the
prediction facility already built into the parsing process.  The left
hand side could be treated by `can end' possibilities, but a better
idea should be to keep within the purview (`remember') previously derived
constituents that involve current words.

There is a phase to be added after detection of a tag or partial parsing error.
Currently processing will just backtrack to the intraword correction level,
but particularly if there has been no correction yet made, {\it pet} should
consider here the possibility of a simple phrase error.  Examples are word
doubling and omission of a common function word.

\bigskip
\noindent
{\it Various Extensions}

\smallskip
Damerau transformations involving the space character (e.g. splitting a word)
have not been implemented yet.  Handling deletion of a space, or substitution
of another character for a space, are straightforward additions to the
morphological
process.  Transposition of a space could be dealt with by setting up an
expectation upon discovering deletion of the last character of a word that the
`deleted' character may be attached to the beginning of the next word.
Addition of a space is trickier because of the focus on the word as a
processing
unit, e.g. corrections for ``the re" could include ``there" or ``the red", but
the present system will not generate the former possibility.

At present the word in focus is always the newest word in the purview.
Altering this would provide some right hand context information, which
would among other things facilitate handling space addition.  Allowing this
change would necessitate a more complex backtracking mechanism, as there would
be a focus lag between morphological processing and later phases.

It would be sensible to keep a reference to the wider context, i.e.
be able to refer to earlier detections/corrections.  With
respect to the editor that {\it pet} is attached to, this could
correspond to a log of errors already encountered in the file being edited.
A recent Microsoft product \footnote{Microsoft Word 6.0 Autocorrect Wizard}
keeps a record of personal habitual
mistakes.  Either could be a valuable aid in choosing the correct correction.

The system could possibly make better use of the graph state of its lexicon.
Word transformation implies either implicit or explicit string comparison.
The advantage of a graph over a trie is that it allows for comparison from
the end of the word and well as the beginning.

\section*{Testing and Evaluation}
With the aim of evaluating the effectiveness of shallow processing,
tests will be carried out
to see what proportion of different types of errors can be dealt with
elegantly, adequately and/or efficiently.  Under examination will be
the number of errors missed/caught and wrongly/rightly corrected.
Different components and
configurations of the system will be compared, for example the
error rules v. p.b.t.'s.  Parameters of the
system will be varied, for example the breadth of the purview, the
position of the purview focus, the number of correction candidates and
the timing of their generation.  Will shallow processing miss too
many of the errors cooperative error processing is aimed at?

There are two significant difficulties with collecting test data.
The central difficulty is finding a representative sample of
genuine errors by native speakers, in context, with the correct version of
the text attached. Apart from anything else, `representative' is
hard to decide - spectrum of errors or distribution of errors ?
Secondly, any corpus of text usually contains only those errors that were
left undetected in the text.  Cooperative processing deals with errors that one
backtracks to catch; if not a different class or range, these at least might
have
a different distribution of error types.

The ideal data would be records of peoples' keystrokes when interacting with
an editor while creating or editing a piece of text.  This would allow
one measure of the (linguistic) feasibility of cooperative error
processing: the effectiveness of shallow processing over errors revealed
by the keystroke-record data.  There does not appear
to be an English source of this kind, so it is planned to compile one.

For comparison, a variety of other data has been collected.  Preliminary tests
used generated errors, from a program that produces random Damerau slips
according
to an observed distribution (Pollock, 1983), using confusion matrices where
appropriate (Kernighan, 1990).  Assembled data includes the
Birkbeck corpus (Mitton, 1986) and multifarious misspelling lists (without
context).  Suggestions have been made to look for low frequency words in
corpora and news/mail archives, and to the Longmans learner corpus
(not native speakers).

\section*{Acknowledgements}
 Thanks to all who offered advice on finding data, and to Doug McIlroy,
Sue Blackwell and Neil Rowe for sending me their misspelling lists.

This work is supported by a British Telecom Scholarship, administered
by the Cambridge Commonwealth Trust in conjunction with the Foreign and
Commonwealth Office.

\end{document}